\documentclass{article}

\usepackage[preprint]{neurips_2019}

\usepackage[utf8]{inputenc} % allow utf-8 input
\usepackage[T1]{fontenc}    % use 8-bit T1 fonts
\usepackage{hyperref}       % hyperlinks
\usepackage{url}            % simple URL typesetting
\usepackage{booktabs}       % professional-quality tables
\usepackage{amsfonts}       % blackboard math symbols
\usepackage{nicefrac}       % compact symbols for 1/2, etc.
\usepackage{microtype}      % microtypography
\usepackage[dvipsnames]{xcolor}
\usepackage{graphicx}
\usepackage{placeins}
\usepackage{float}
\usepackage{soul}

\title{Formatting Instructions For NeurIPS 2019}
\usepackage{amscd}
\usepackage{amsmath} 
\usepackage{cleveref}
\usepackage[format=hang,singlelinecheck=0,font={sf,small},labelfont=bf,caption=false]{subfig}
\usepackage{caption}
\usepackage{afterpage}

\newcommand{\vect}[1]{#1} 
\newcommand{\vwone}{\vect{w}^{(1)}} 
\newcommand{\vwtwo}{\vect{w}^{(2)}}
\newcommand{\vwonehat}{\hat{\vect{w}}^{(1)}}

\newcommand{\act}[1]{\phi \left( #1 \right) }

\Crefname{equation}{Eq.}{Eqs.}
\Crefname{figure}{Fig.}{Figs.}
\Crefname{tabular}{Tab.}{Tabs.}
\Crefname{section}{Sec.}{Secs.}
\creflabelformat{equation}{#2#1#3}

\title{Dimensionality compression and expansion in Deep Neural Networks}

\author{%
Stefano Recanatesi\thanks{These authors contributed equally.}\\ Center for Computational Neuroscience\\ University of Washington\\ Seattle, WA\\
\texttt{stefanor@uw.edu} \\
   \And
   Matthew Farrell$^*$ \\
   Center for Computational Neuroscience \\
   University of Washington \\
   Seattle, WA \\
   \texttt{msf9@uw.edu} \\
   \AND
   Madhu Advani \\
   Center for Brain Science \\
   Harvard University \\
   Cambridge, MA \\
   \texttt{madvani@fas.harvard.edu} \\
   \And
   Timothy Moore \\
   Center for Computational Neuroscience \\
   University of Washington \\
   Seattle, WA \\
   \texttt{tjm36@uw.edu} \\
   \AND
   Guillaume Lajoie \\
   Dept. of Mathematics and Statistics \\
   Universit\'e de Montr\'eal \\
   Montr\'eal, Qu\'ebec, Canada \\
   \texttt{lajoie@dms.umontreal.ca} \\
   \And
   Eric Shea-Brown \\
   Center for Computational Neuroscience \\
   University of Washington \\
   Seattle, WA \\
   \texttt{etsb@uw.edu} \\
}

\begin{document}

\maketitle

\begin{abstract}
Datasets such as images, text, or movies are embedded in high-dimensional spaces.  However, in important cases such as images of objects, the statistical structure in the data constrains samples to a manifold of dramatically lower dimensionality.  Learning to identify and extract task-relevant variables from this embedded manifold is crucial when dealing with high-dimensional problems. We find that neural networks are often very effective at solving this task and investigate why. To this end, we apply state-of-the-art techniques for intrinsic dimensionality estimation to show that neural networks learn low-dimensional manifolds in two phases:  first, dimensionality expansion driven by feature generation in initial layers, and second, dimensionality compression driven by the selection of task-relevant features in later layers.  Our mathematical analysis shows how Stochastic Gradient Decent balances the dimensionality of neural representations by inducing an effective regularization term in the loss. We highlight the important relationship between low-dimensional compressed representations and generalization properties of the network. Our work contributes by shedding light on the success of deep neural networks in disentangling data in high-dimensional space while achieving good generalization. Furthermore, it invites new learning strategies focused on optimizing measurable geometric properties of learned representations, beginning with their intrinsic dimensionality.
\end{abstract}

\section{Introduction}
Deep neural networks are able to accurately classify high-dimensional data, not only achieving high training accuracy but also generalizing well to held-out samples. This is in spite of the myriad challenges associated with high-dimensional spaces, often referred to collectively as the \emph{curse of dimensionality}. This is also in spite of deep networks typically existing in a highly over-parameterized regime where the number of parameters greatly exceeds the number of data samples. What is the reason for this unreasonable effectiveness? Here, we find new answers by probing networks with nonlinear metrics for dimensionality and developing theory that shows how deep networks naturally learn to compress the representation dimensionality of their inputs, sidestepping its apparent curse.

For concreteness we consider image classification datasets, but the observations and arguments we make are more general.  While a dataset of images is naturally embedded in a high-dimensional space -- the rgb space of 32x32-pixel images has dimension 3072 -- the statistics of the dataset generally constrain the images to  lie on a lower-dimensional structure, a nonlinear \emph{manifold} \citep{Edraki_2018_ECCV}.  
What is the shape and dimension of this manifold, and how does learning influence these attributes? Recent developments in data science have yielded techniques for estimating the intrinsic dimensionality of manifolds which are robust to the high dimensionality of the embedding space as long as the manifold itself is low-dimensional \citep{hinton_reducing_2006,van_der_maaten_dimensionality_2009, campadelli_intrinsic_2015,camastra_intrinsic_2016}. Here we deploy these state-of-the-art tools to analyze the dimensionality of image datasets (Fashion-MNIST and CIFAR-10) and of their deep manifold representations.

We train two deep neural network models to classify images from these datasets, and we use local and global metrics for dimensionality -- the first time they have been applied to deep neural networks to the best of our knowledge -- to analyze the geometry of the resulting manifold representations at each layer through the network architecture. Throughout training, each layer develops a specific  representation in its high-dimensional neural space with properties determined by both task demands and by learning mechanisms. We find:
\begin{enumerate}
\item  The dimensionality of representation manifolds is very low when compared to the number of neurons in each layer. 

\item The dimensionality of representation manifolds evolves through trained networks in two distinct phases: initial layers expand dimensionality, and final layers compress it~\citep{Guyon2003AnIT,vincent_extracting_2008,wang2014role,tishby_information_2018}. 

\item Dimensionality expansion and compression are automatically balanced by SGD, and can be understood through an effective loss with two competing terms: one enforcing task demands on training data, and one compressing the manifold dimension.

\end{enumerate}

Our results on the low dimensionality of learned deep representation manifolds helps explain why deep networks show good generalization properties despite using massive numbers of parameters: low-dimensional or otherwise minimal representations are thought to support good generalization~\citep{fusi_why_2016,zhang2016understanding,tishby_information_2018}.  We close the paper by discussing how probing and controlling dimensionality suggests new avenues to improve both AI and our understanding of neural coding strategies in brain circuits.

\begin{figure*}[htb!]
\subfloat{\label{fig:1a}}
\subfloat{\label{fig:1b}}
\subfloat{\label{fig:1c}}
\subfloat{\label{fig:1d}}
\subfloat{\label{fig:1e}}
\subfloat{\label{fig:1f}}
\centering
\includegraphics[width=1.\textwidth]{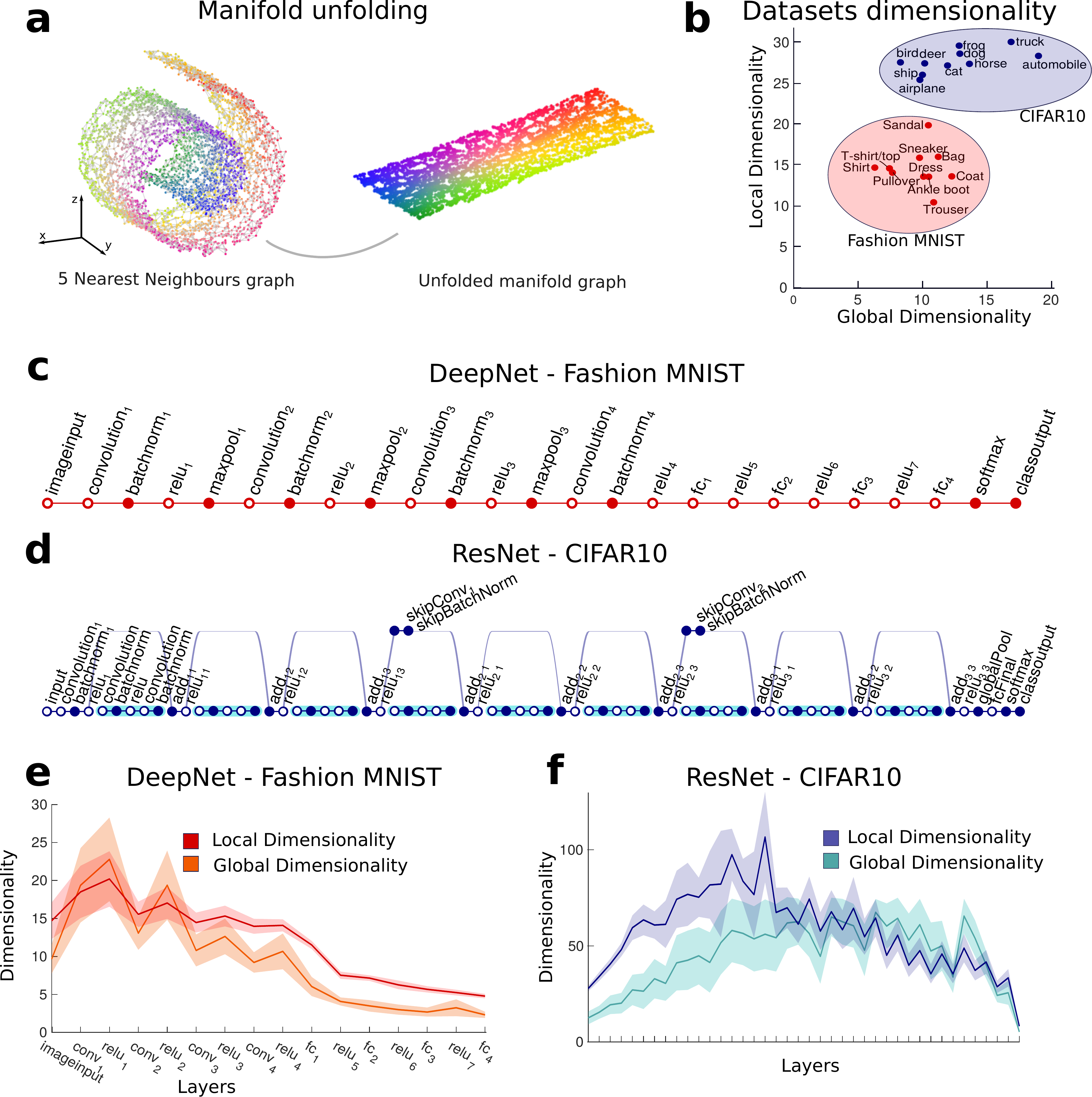}
\centering
\caption{Dimensionality analysis of representation manifolds. \protect\subref{fig:1a}) Example of unfolding a Swiss-roll manifold. \protect\subref{fig:1b}) Local and Global dimensionality of data manifolds for Fashion-MNIST and CIFAR-10. \protect\subref{fig:1c}) Structure of DeepNet for classifying Fashion-MNIST. \protect\subref{fig:1d}) Structure of ResNet for classifying CIFAR-10. \protect\subref{fig:1e}) Local and global dimensionality of manifold representations through the layers of DeepNet and ResNet (\protect\subref{fig:1f}). Error bars indicate 95\% confidence intervals across classes. The displayed layers are the ones colored white in panels (\protect\subref{fig:1c}) and (\protect\subref{fig:1d}).}
\label{fig:1}
\end{figure*}

\section{Methodology}
\paragraph{Intrinsic dimensionality and its estimation}
Estimating the dimensionality of data manifolds is crucial for understanding how and why neural networks learn -- but it is difficult, because linear analyses often fail to capture the effect of nonlinearities in embedding low-dimensional structures in high-dimensional spaces.  By employing novel techniques \citep{granata_accurate_2016,facco_estimating_2017} we overcome the limits of previous analyses based on linear methods \citep{zhang_local_2017}, uncovering new important phenomena.
This builds on a rich literature on the estimation of intrinsic dimensionality of manifolds ~\citep{grassberger_measuring_1983,tenenbaum_global_2000,costa_manifold_2003,levina_maximum_2005,van_der_maaten_dimensionality_2009,campadelli_intrinsic_2015,camastra_intrinsic_2016}.  Here we provide a brief treatment of the essential concepts of intrinsic dimensionality estimation.

By means of example consider a simple sheet of paper. On a \emph{local scale} a paper has three dimensions on which its molecules are organized, although on a more \emph{global scale} we could say that it has only two dimensions on which we may draw or print. Importantly, independently of how it is folded or crumpled, these properties are persistent: locally it is three dimensional, globally it is two dimensional. To formalize these ideas we consider our sheet of paper folded to resemble a Swiss-roll~\citep{silva2003global}, \Cref{fig:1a}, this is a curved manifold with local dimensionality of 3 and global dimensionality of 2 embedded in 3d. We remark that embedding the paper in a four or higher dimensional space would leave its local and global dimensionality unaffected; this point is very important in the following where we consider the local and global properties of manifolds embedded in high-dimensional spaces.

In the example above the dimensionality changes as a function of the radius -- this property is called multiscaling, a common intrinsic property of statistical manifolds ~\citep{silva2003global,camastra_intrinsic_2016,little_multiscale_2017}.
For this reason we study two different measures of dimensionality, at a local scale ($r \rightarrow 0$)~\citep{facco_estimating_2017} and a global scale (for values of $r$ around the mode value)~\citep{granata_accurate_2016}. In the local case the dimensionality is computed from the scaling of the probability distribution of nearest neighbor distances $\mathcal{P}_{NN}(r)$. A linear fit of the log probability of this distribution is proportional to the intrinsic dimensionality $d$. The 95\% confidence interval on the fit is used to report uncertainty (cf.~\citep{facco_estimating_2017} for details).
In the global case the k-nearest-neighbor graph ($k=20$) is built and geodesic distances are computed. The resulting probability distribution of global distances $\mathcal{P}_{G}(r)$ is then analyzed around its mode value $r_m$. Specifically the portion of the distribution falling in between $r_m-r_{\sigma}$ and $r_m+\frac{1}{2}r_{\sigma}$, where $r_{\sigma}$ is the standard deviation of the distribution, is compared to the same portion of the distribution of distances between points drawn from a hypersphere of varying dimensionality $d$. The value of $d$ which minimizes the least square error between the two portions of distributions in the considered interval is the estimated global intrinsic dimensionality (cf.~\citep{granata_accurate_2016} for details). These two methods have been chosen on a criterion of robustness and minimality.

Related intrinsic dimensionality estimation methods \citep{costa_learning_2004,levina_maximum_2005,ceruti_danco:_2012} yield consistent results with the metrics here selected whenever a robust convergence is achieved. 
In \Cref{fig:1b} we visualize the local and global dimensionality for the training set of the ten classes of CIFAR-10 and Fashion-MNIST, where the dimensionality of each class is measured individually. Different classes have slightly different dimensionalities but are overall consistent in their value of global and local dimensionality. This suggests that the methods we use are able to extract consistent information from datasets with similar statistics. Linear techniques based on singular value decomposition or principal component analysis are not able to provide such an accurate dimensionality estimation, largely overestimating the dataset dimensionality (data not shown). An alternative approach to measuring dimensionality of representation manifolds was developed in \cite{chung_2016,chung_classification_2018} and recently applied to deep neural networks in \cite{cohen_separability_2019}. This measure captures the arrangement of class manifolds in space from the perspective of a maximum margin linear classifier.

\paragraph{Deep network representation spaces}
Next we turn to the intrinsic dimensionality of the representations developed in feedforward neural networks. To assess the dimensionality of deep representations~\citep{bengio_representation_2013} we considered two benchmarks: a deep neural network trained to classify Fashion-MNIST and a ResNet~\citep{szegedy2017inception} trained to classify both CIFAR-10 and CIFAR-100. The architectures of the two networks are reported respectively in \Cref{fig:1c} and \Cref{fig:1d}. The two networks were trained with SGD with a starting learning rate of 0.01 decreasing linearly by 0.0001 per epoch (fixed policy). Both networks were trained for 100 epochs and then the epoch where the validation accuracy was first minimized within a 0.1\% accuracy was selected. The two networks achieved 90.96\% and 87.6\% testing accuracy, respectively. Importantly network architectures were chosen that keep the layer width constant as much as possible. DeepNet layer sizes decreased only via max-pooling layers. The 784 input variables passed through the architecture in \Cref{fig:1c}, where the total layer size decreases at each occurrence of a max-pooling or fully connected layer according to the sequence (25088, 6272, 1568, 288, 64, 10). Similarly in ResNet the initial 3072 variables followed the sequence of layer sizes (16384, 8192, 4096, 64, 10) throughout the network, decreasing only in the the case of Skip Convolution or fully connected layers. This helps disentangle the effect of layer sizes on representation dimensionality.
Each layer induces a set of representation manifolds over the ensemble of inputs, one manifold corresponding to each class. To measure intrinsic dimensionality, we compute the dimensionality of the representation manifold for each class individually and average the results, reporting the 95\% confidence intervals of the deviations.

\section{Intrinsic dimensionality of learned representations}
We computed the intrinsic dimensionality of deep representations for two deep neural networks, DeepNet and ResNet, trained on the Fashion-MNIST and CIFAR-10 datasets, respectively. Initial layers expanded the dimensionality of the input dataset while final ones carried out a dimensionality reduction (cf. \Cref{fig:1e} and \Cref{fig:1f}).

\Cref{fig:1e,fig:1f} indicate roles for specific layer types in increasing and decreasing dimensionality. In particular, ReLU nonlinearities consistently increased the dimensionality of their inputs by a factor that we measure to be $1.16\pm 0.19$ across all network instances and classes (data not shown).
Dimensionality compression was driven primarily by the application of the weight matrix before the ReLU nonlinearity was applied.
Early convolutional layers tended to increase dimensionality.
This highlights the tools that the network can use to create higher-dimensional as well as lower-dimensional feature representations.

Note that dimensionality expansion is related to feature generation ~\citep{olshausen_sparse_1997,babadi2014sparseness}, akin to support vector machine kernel space expansion, and dimensionality reduction can be viewed in terms of feature selection~\citep{hinton_reducing_2006,tishby_information_2018}. In the context of image classification, while earlier layers are often thought of as generating spatial features, final layers are thought to select and combine features that are relevant to classifying images according to their labels~\citep{wang2014role}.

We emphasize that estimation properties make computing intrinsic dimension in high-dimensional spaces a serious challenge.  A common approach in this setting is to consider multiple metrics. Here, both dimensionality metrics plotted above -- and others not shown -- are in agreement with the trends we describe. This strengthens our confidence in identifying robust trends of the dimension of representation manifolds.

\begin{figure*}[htb!]
\subfloat{\label{fig:2a}}
\subfloat{\label{fig:2b}}
\subfloat{\label{fig:2c}}
\subfloat{\label{fig:2d}}
\centering
\includegraphics[width=1.\textwidth]{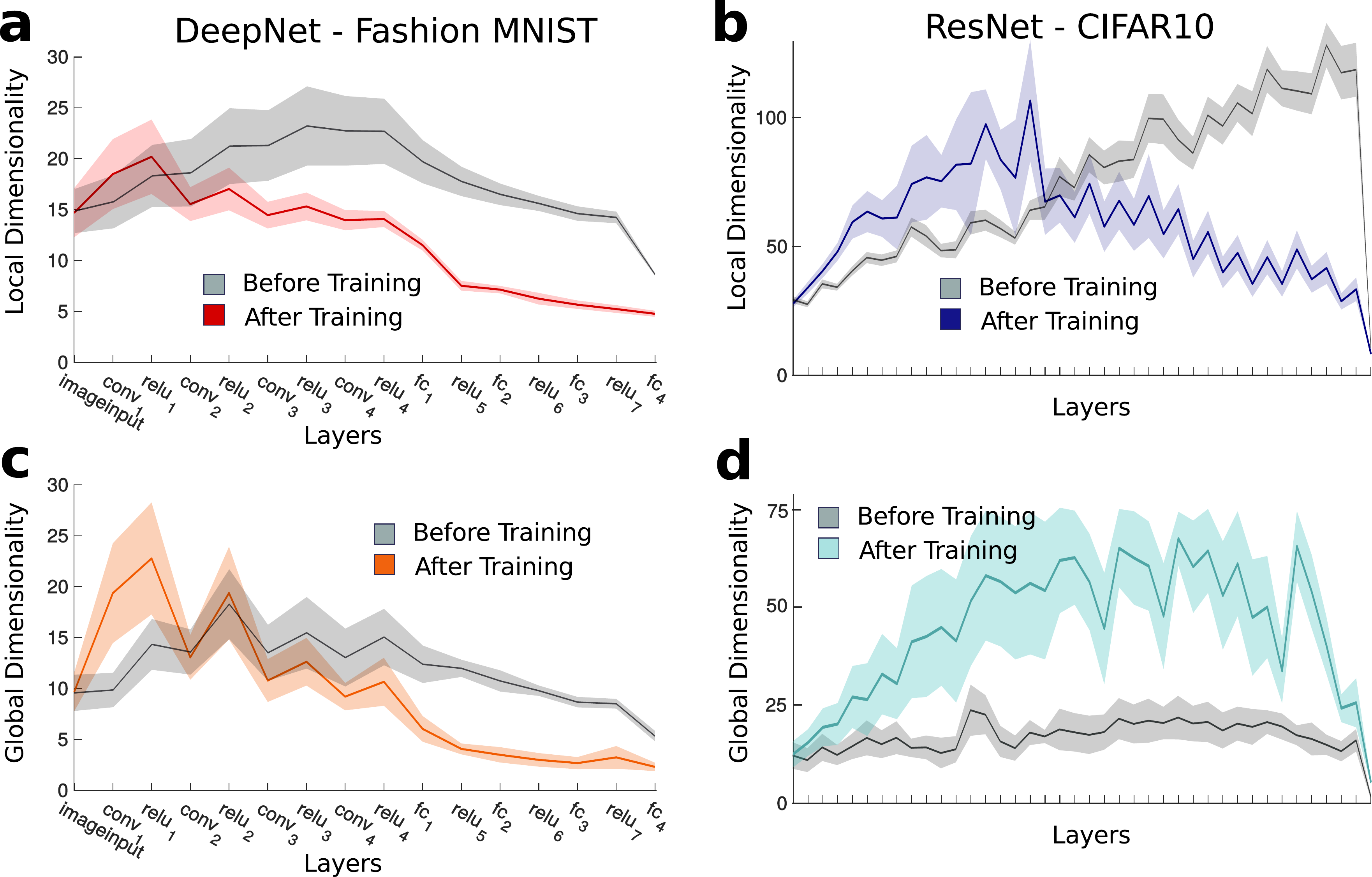}
\centering
\caption{Dimensionality of representation manifolds before and after training. \protect\subref{fig:2a}) Local dimensionality of manifolds through the layers of DeepNet before and after training. Global dimensionality analysis as in (\protect\subref{fig:2c}). \protect\subref{fig:2b}) Local dimensionality for ResNet before and after training. Global dimensionality analysis as in (\protect\subref{fig:2d}). Error bars indicate 95\% confidence intervals across class manifolds.}
\label{fig:2}
\end{figure*}

\section{The role of learning in shaping the dimensionality of representations}
How does training shape the dimensionality of network representations?
To address this question we compare local and global dimensionality before and after training, for DeepNet (\Cref{fig:2c}) and ResNet (\Cref{fig:2a,fig:2b}). 

Compared with the untrained network, training slightly increased local dimensionality in initial layers, and significantly decreased it in final ones (\Crefrange{fig:2a}{fig:2b}).  We note that before training, both DeepNet and ResNet showed the same layer-specific effects for local dimensionality: convolutional layers tended to expand local dimensionality while fully connected layers tended to decrease it (\Cref{fig:2a} and \Cref{fig:2b}). ResNet -- a network that has convolutional layers throughout its depth -- exhibited this phenomenon most clearly, with a nearly monotonic increase in dimension before training.
For DeepNet, training had the same effects on global dimensionality as for local dimensionality (\cref{fig:2c}), increasing these dimensionalities in early layers and decreasing them in later ones.  However, for the ResNet trained on CIFAR-10 (\cref{fig:2d}), learning increased the global dimensionality of all the layers, while the increasing-decreasing trend across layers was preserved after learning. 
We hypothesize that this occurs because ResNet primarily extracts local features before learning, expressing them globally only after learning.
This is evidenced by the significant difference between local and global dimensionalities for the untrained networks  (\Cref{fig:2b} vs. \Cref{fig:2d}).  Training serves to express local information useful for solving the task globally; once that has occurred, local and global dimensions become nearly equal (\Cref{fig:1f}).

Overall the results shown here (\Crefrange{fig:2a}{fig:2d}) are consistent with the interpretation that trained neural networks generate high-dimensional collections of features in early layers and select out a low-dimensional combination of these features in later layers.
This suggests that the network is driven both by the need to expand and by the need to compress the dimensionality of its representation, and that the learned behavior of the network constitutes a balancing of these two demands.
Which mechanisms induce the network to strike this optimal balance? In the next section we show how SGD itself naturally produces these two complementary effects.

\begin{figure*}
%\captionsetup{justification=justified}
\subfloat{\label{fig:3a}}
\subfloat{\label{fig:3b}}
\subfloat{\label{fig:3c}}
\subfloat{\label{fig:3d}}
\centering
\includegraphics[width=1.\textwidth]{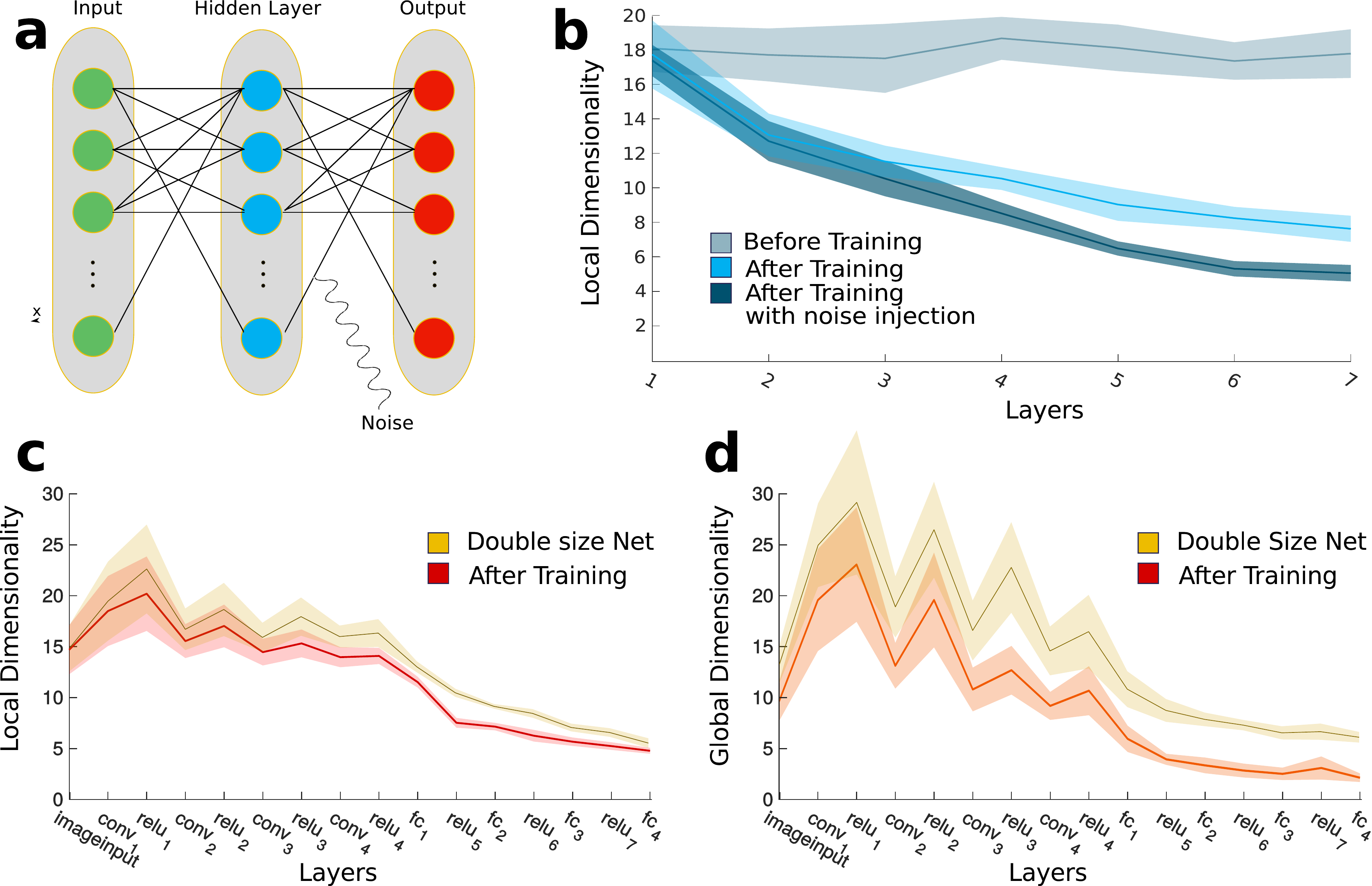}
\centering
\caption{Theoretical analysis of SGD dimensionality compression. \protect\subref{fig:3a}) Schematics of a 2 layer networks with noise injected in the second layer.   \protect\subref{fig:3b}). Dimensionality of deep network with constant layer size of 200 trained to classify Fashion-MNIST. The three lines display the dimensionality before and after training with and without the injection of noise in all the feedforward connections ($\sigma=0.005$). Input weights are initialized to be random while all other weights are initialized to the identity matrix. \protect\subref{fig:2c}) Local dimensionality of DeepNet representation manifolds after training for the original network and the same network with intermediate layers doubled in size. Same for global dimensionality in panel (\protect\subref{fig:2d}). Error bars indicate two standard deviations.}
\label{fig:3}
\end{figure*}

\section{SGD balances task demands with dimensionality compression}
We analyze a two-layer neural network trained to classify inputs according to $c$ classes (see \Cref{fig:3a}). The equations for the network are:
\begin{align}
\begin{split}
	& h_{i} = \sum_{j} \act {\vwone_{ij} x_{j}}\\
	& \hat{y}_{i} = \sum_{j}{\vwtwo_{ij} h_j},\
\end{split}
\end{align}
where $x, h$ and $\hat{y}$ are the input, the hidden representation, and the output of the network, respectively. Additionally, $\vwone, \vwtwo, \phi$ are respectively the input weights, readout weights, and a (possibly nonlinear) activation function. We consider the cost function $L(\Theta) = \sum_{\mu=1}^P \left\Vert y^\mu - \hat{y}^{\mu}\right\Vert_{2}^{2}$ where $\Theta = [\vwone,\vwtwo]$ and $\mu$ indexes the training set of size $P$. Here the output $\hat{y}^{\mu}$ is a length $c$ vector, and the targets $y^{\mu}$ one-hot encode the labels.

Over training, SGD generates effective noise in the parameter updates (see e.g. \citep{zhang_energy-entropy_2018, smith2017bayesian}) due to the fact that each update is performed on a subset of the training data. In general SGD leads to noisy gradient updates of the form:
\begin{equation}
    \Delta \Theta = -\eta \nabla_\Theta L(\Theta) + z_t
\end{equation}
where $\eta$ is the learning rate and $z_t$ is the noise generated from each mini-batch, or the difference between the gradient of the full batch and mini-batch $t$.
This noise in the gradient updates is correlated. Here we simplify the analysis by modeling this noise in parameter updates by adding noise of variance $\sigma^2$ directly to the output weights $\vwtwo$, and by assuming this noise is isotropic Gaussian with zero mean.
This is the simplest setting sufficient to provide intuition for the phenomena highlighted in this work: namely expansion and compression of the dimension of representations.
Our analysis leads to the effective cost function:
\begin{equation} \label{eqn:loss} 
	L(\Theta, \sigma) = \sum_{\mu=1}^P \sum_{i}( y_{i}^\mu - \sum_{j}(\vwtwo_{ij} +\sigma \xi_{ij}^{\mu}) h_{j}^{\mu} )^2,
\end{equation}
where $\xi$ is Gaussian noise with unit variance independently sampled across $\mu$. Taking an average over $\xi$, for a learning rate small enough, we can rewrite \Cref{eqn:loss} in the form:
\begin{equation}\label{eqn:eff_cost}
     L(\Theta,\sigma)\approx \langle L(\Theta,\sigma)\rangle_{\xi} = L(\Theta,0) + R(\sigma) = L(\Theta,0) + \sigma^2 \text{Tr}(C),
\end{equation}
with $C_{j j'} = \sum_\mu{ h^\mu_j h^\mu_{j'}}$ and where we view $R(\sigma)=\sigma^2 \text{Tr}(C)$ as a regularization term. Similar effective regularization terms have been shown to arise from dropout \citep{wager2013dropout,Goodfellow-et-al-2016}, and the effects of regularization on generalization have been heavily studied \citep{Goodfellow-et-al-2016}. Here we focus on the effects of such regularization in shaping the dimensionality of the representation.
While the compressive effects of an initial step of SGD have been previously noted \citep{Farrell564476}, here we consider the limit of many steps. 

Geometrically, all the directions of the representation $\vect{h}^{\mu}$ in the span of the readouts $\vwtwo$ contribute to the cost both in $L(\Theta,0)$ and $R(\sigma)$, while the directions orthogonal to this span contribute only to the regularization penalty $R(\sigma)$.
By penalizing the norm of the representation $\Vert h^{\mu}\Vert^{2}_{2}$, the regularizer $R(\sigma)$ encourages the reduction of all $h^{\mu}$ components, including the orthogonal components $\vect{h}_{\perp}^{\mu}$ of the representation with respect to the readout weights $\vwtwo$.
We refer to this action as compression of task-irrelevant directions (directions orthogonal to the readout $\vwtwo$). For $\sigma > 0$ there is indeed a unique solution  $(h^{\mu})^{*}$ with null orthogonal components $(\vect{h}^{\mu}_{\perp})^{*}=0$ that minimizes the cost \cref{eqn:eff_cost}:
$(h^{\mu})^{*} = ((\vwtwo)^{T}\vwtwo + \sigma^2 I)^{\dagger} (\vwtwo)^{T} y^{\mu}$. Here $\dagger$ denotes the Moore-Penrose pseudo-inverse.
The uniqueness is a consequence of the strict convexity of $R(\sigma)$ and the convexity of $L(\Theta,0)$ as functions of $h^{\mu}$ when $h^{\mu}$ is unconstrained. The cost increase of straying from this solution is quadratic.

A network that learns the task balances minimizing the task cost $L(\Theta,0)$ with encouraging the compression of task-irrelevant directions due to $R(\sigma)$. For instance, learning a  task that is not linearly separable with large amounts of training data requires a higher-dimensional hidden representation, which may come at the expense of increasing $R(\sigma)$, since $R(\sigma)$ increases isotropically as $h^{\mu}$ diverges from $(h^{\mu})^{*}$. In other words, balancing the two terms $L(\Theta,0)$ and $R(\sigma)$ in the loss shapes the representation $h^{\mu}$ so that the manifold dimension of $h^{\mu}$ is expanded only when aiding the reduction of the task loss $L(\Theta,0)$ by separating the classes (see \citep{Cover4038449} for formal connections between dimensionality expansion and class separation).

To provide further intuition regarding this balance, we consider the case of linear activations $\left( \phi(z) = z \right)$. In this setting we can write a closed form expression for the first layer weights. When $\sigma > 0$ the cost \Cref{eqn:eff_cost} is strictly convex with respect to the weights, and the unique minimizer $\vwonehat$ is:
\begin{equation}
\vwonehat = ((\vwtwo)^{T} \vwtwo + \sigma^2 I)^{\dagger}(\vwtwo)^{T} Y^{T}X (X^{T} X)^{\dagger},
\end{equation}
where $X$ is a $P \times d$ matrix of input samples and $Y$ is a $P\times c$ matrix of labels. Here $d$ is the embedding space dimension for the inputs. This equation reveals that the range of $\vwonehat$ lies within the span of the output weights, which implies that $\hat{h}^{\mu} = \vwonehat x^{\mu}$ lies in the span of the output weights as well. Equivalently, $\hat{h}_{\perp}^{\mu} = 0$ for all $\mu$.  Strict convexity assures that for an appropriate learning rate scheme, SGD will converge to this solution. The linear network is nearly always able to achieve $(h^{\mu})^{*}$ when $P\le d$,
but not necessarily for larger numbers of samples. However, in either setting it will still remove all task-irrelevant directions from the representation.

We can consider how well these results generalize to the commonly applied ReLU nonlinearity ($\phi(x) = \max\{x, 0\}$). In the limit as $\sigma \rightarrow 0$, the hidden activity minimizing the loss will have the form $h^* + h_\perp$, where $w^{(2)} h_\perp = 0$. The addition of $h_\perp$ will not impact the ability to fit the training data, but is required to satisfy the nonnegativity constraint imposed by the ReLU activation. Thus, the dimensionality of the representation is larger in this nonlinear setting, and as the regularization strength is increased it leads to a trade-off between reducing dimensionality and fitting the training data.

Our theory suggests that the effective regularizer $R(\sigma)$ induced by fluctuations in weight updates encourages the compression of task-irrelevant directions and that dimensionality expansion of the activity in the hidden layer can occur when task complexity requires it. Although these results have been developed in the context of a two-layer network we hypothesize that they apply more broadly to deep networks. In this case the noise in the ``output'' weights of each layer generated by SGD encourages compression of the hidden representation of each layer.
To support this conjecture we track the dimensionality of the ten Fashion-MNIST classes through a 7 layer feedforward fully connected network with 200 units per layer and a ReLU nonlinearity in \Cref{fig:3b}.
Upon training with SGD on a mean squared error loss, we see that the dimensionality reduction through layers is more pronounced when noise is injected into the weights during the training process (dark blue line). This is what the theory above predicts, as higher effective noise induces a larger regularization $R(\sigma)$ and in turn stronger compression and lower dimensionality. As also predicted, we found a similar trend for another manipulation that increases the effective noise, decreasing
the batch size, with smaller batches inducing stronger compression (data not shown).

Finally, we note that the learning of a low-dimensional representation is thought to prevent overfitting and aid generalization.
Formal bounds on generalization are typically written in terms of the complexity of the class of functions used to fit data or of the parameters learned by the network (such as in ~\citep{Liang:2017vv,Vapnik1998}).
To the best of our knowledge, formal theoretical links between dimensionality of deep neural network representations and generalization have not been explicitly established.
However, it is intuitive that representation geometry \citep{shao2018riemannian,shukla2019geometry} is connected to these ideas.
Lower-dimensional distributions require fewer samples be drawn before the structure of the distribution can be inferred \citep{MR2298361,fusi_why_2016}.
This means that weights trained to transform a low-dimensional representation will require fewer training samples before the true distribution is learned \citep{fefferman2015reconstruction,fefferman2016testing}, i.e. before the weights generalize.
See \citep{fusi_why_2016,rigotti_importance_2013,zhang2016understanding,tishby_information_2018} for more in-depth discussions of this topic.

Our simulations and theoretical analysis suggest that the dimensionality of the representations in deep networks is driven by balancing training data task demands with compression of task-irrelevant directions, as opposed to being driven by the number of neurons in the layers.
To provide evidence for this argument, we double the number of units in each layer of DeepNet before training and compare
the dimensionality of its representations (\Cref{fig:3d}). The accuracy (which increased by 2.1\%, data not shown) and the dimensionality of layer representations are only slightly affected, despite the major increase in model complexity. Similarly, training ResNet on CIFAR-100 in place of CIFAR-10 doesn't lead to a significant increase in the layers' manifold dimensionality (data not shown). This supports the hypothesis that the SGD learning rule discovers a minimal-dimensional manifold that solves the task, and does so in a way that is insensitive to network size -- helping to make the generalization power of deep architectures robust to over-parametrization.

\section{Conclusion}
In this paper we deploy state-of-the-art nonlinear dimensionality estimation techniques to measure the intrinsic dimensionality of deep neural networks' activations during a classification task. Our results show that the representation manifolds are very low-dimensional when compared to the network architecture, on the order of ten dimensions compared to the thousands of neurons per layer. We identify two distinct phases of dimensionality expansion and compression through the network's layers.
A natural interpretation is that the expansion phase generates features that aid in solving the task \citep{babadi2014sparseness,fusi_why_2016}, while the compression phase selects out the key task-relevant features from among those generated \citep{wang2014role}.
As an example, feature selection allows for the building of invariances to class-irrelevant transformations (such as rotations and translations).
Both simulation and theoretical analysis of SGD demonstrate that these phases emerge as learning balances task demands with a tendency to compress the manifold dimension. See \citep{cohen_separability_2019} for recent work that finds similar trends for a different measure of dimensionality.

Our advances in measuring and understanding trends in the dimensionality of representation manifolds have a number of applications. The minimal number of neurons needed in the bottleneck of an autoencoder to reproduce the dataset can be viewed as an alternative measure of intrinsic dimensionality of the dataset (by this measure MNIST is believed to has dimensionality $\sim$13) ~\citep{hinton_reducing_2006,yu2018understanding,camastra_intrinsic_2016}.
Here we move beyond considering the intrinsic dimensionality of datasets to study the dimension of deep representations of these datasets that networks use to classify them. Following the autoencoder example, we posit that our results may provide a foundation for future work to determine the most efficient sizes of networks that learn classification tasks
\citep{han2015deep,tu2016reducing,tung2018deep}. For instance, if the maximum dimensionality achieved by a network is 50 in a middle layer, we conjecture that this will inform the layer size of a deep neural network that solves the task with high performance, either via standard training procedures or those that add pruning or compression steps.

Furthermore, our analysis suggests that representations learned by deep architectures tend to be low-dimensional due to the intrinsic regularization properties of SGD.  Analyzing terms in an approximate loss shows that SGD encourages the representation manifold to be as low-dimensional as possible without compromising task accuracy, in effect removing task-irrelevant dimensions.  We note that ~\cite{tishby_information_2018} and preceding works advanced similar ideas in terms of task-irrelevant information.
Overall, low-dimensional representations add perspective to the community's efforts to link deep neural networks to generalization, ~\citep{fusi_why_2016,zhang2016understanding,tishby_information_2018,advani_high-dimensional_2017,wu2017towards}, and will likely be critical in furthering understanding of why deep networks generalize well ~\citep{novak_sensitivity_2018}. Moreover, the dimensionality-oriented perspective outlined in this work opens new possibilities in explicitly regularizing networks to improve performance on unobserved data.
Examples include explicitly encouraging dimensionality compression through addition of tailored noise during training, adding dimension regularizing norms explicitly, or encouraging dimensionality expansion through other means including the choice of architecture in early layers. In addition, the connections drawn between noise, dimensionality compression, and generalization provide hints for understanding the representations formed in biological learning circuits \citep{fusi_why_2016}, which are themselves noisy and often low-dimensional \citep{cunningham_dimensionality_2014}.

\section*{Acknowledgements}
This work was supported by NSF DMS Grants \#1514743 and \#1256082, an NSERC Discovery Grant (RGPIN-2018-04821), an FRQNT Young Investigator Startup Program (2019-NC-253251), and an FRQS Research Scholar Award, Junior 1 (LAJGU0401-253188), as well as the Boeing Endowment at University of Washington Applied Mathematics.

\newpage
\small{

\bibliographystyle{plain}
\bibliography{bibpaper.bib}
}
\end{document}